\pdfoutput=1
\newcommand\nnfootnote[1]{%
  \begin{NoHyper}
  \renewcommand\thefootnote{}\footnote{#1}%
  \addtocounter{footnote}{-1}%
  \end{NoHyper}
}

\documentclass[11pt]{article}

\usepackage[]{acl}
\usepackage{times}
\usepackage{latexsym}
\usepackage{graphicx}

\usepackage{subfig}
\usepackage{url}

\usepackage[T1]{fontenc}
\usepackage[utf8]{inputenc}

\usepackage{microtype}
\usepackage{tabularx} 

\usepackage{multirow}
\usepackage{graphicx}
\usepackage{tabularx, multirow, booktabs}
\title{JarviX: A LLM No code Platform for Tabular Data Analysis and Optimization}

\author{\textbf{Shang-Ching Liu}$^{1,3,\dagger}$, \textbf{ShengKun Wang}$^{2,3,\dagger}$,  \textbf{Wenqi Lin}$^3$, \textbf{Chung-Wei Hsiung}$^3$,\\ \textbf{Yi-Chen Hsieh}$^3$, \textbf{Yu-Ping Cheng}$^3$, \textbf{Sian-Hong Luo}$^3$,\textbf{Tsungyao Chang}$^3$, \textbf{Jianwei Zhang*}$^1$ \\
$^1${Department of Computer Science, University of Hamburg} \\
$^2${Department of Computer Science, Virginia Tech} \\
$^3${Synergies Intelligent Systems, Inc.} \\
\texttt{jianwei.zhang@uni-hamburg.de} \\
}

\begin{document}
\maketitle
\nnfootnote{$^\dagger$These authors contributed equally to this work.}
\renewcommand\thefootnote{\arabic{footnote}} 
% \author{
% Shang-Ching Liu \\ Department of Computer Science,\\ University of Hamburg \\
%   \texttt{shang-ching.liu@uni-hamburg.de} \\\And
  % Mathew \\
  % Synergies Research \\
  % \texttt{mathew@sis.ai} \\
  % Euna \\
  % Synergies R\&D \\
  % \texttt{euna.hsieh@sis.ai} \\\And
  % Yu-Ping Cheng \\
  % Synergies R\&D \\
  % \texttt{debbie.cheng@sis.ai} \\\And
  %  Sian-Hong Luo\\
  % Synergies R\&D \\
  % \texttt{andy.luo@sis.ai} 
  
  % }

\begin{abstract}
%\footnotetext[1]{Corresponding author}
In this study, we introduce JarviX, a sophisticated data analytics framework. JarviX is designed to employ Large Language Models (LLMs) to facilitate an automated guide and execute high-precision data analyzes on tabular datasets. This framework emphasizes the significance of varying column types, capitalizing on state-of-the-art LLMs to generate concise data insight summaries, propose relevant analysis inquiries, visualize data effectively, and provide comprehensive explanations for results drawn from an extensive data analysis pipeline. Moreover, JarviX incorporates an automated machine learning (AutoML) pipeline for predictive modeling. This integration forms a comprehensive and automated optimization cycle, which proves particularly advantageous for optimizing machine configuration. The efficacy and adaptability of JarviX are substantiated through a series of practical use case studies.
\end{abstract}

\begin{figure*}[htb!]
    \centering
    \includegraphics[width=0.9\textwidth]{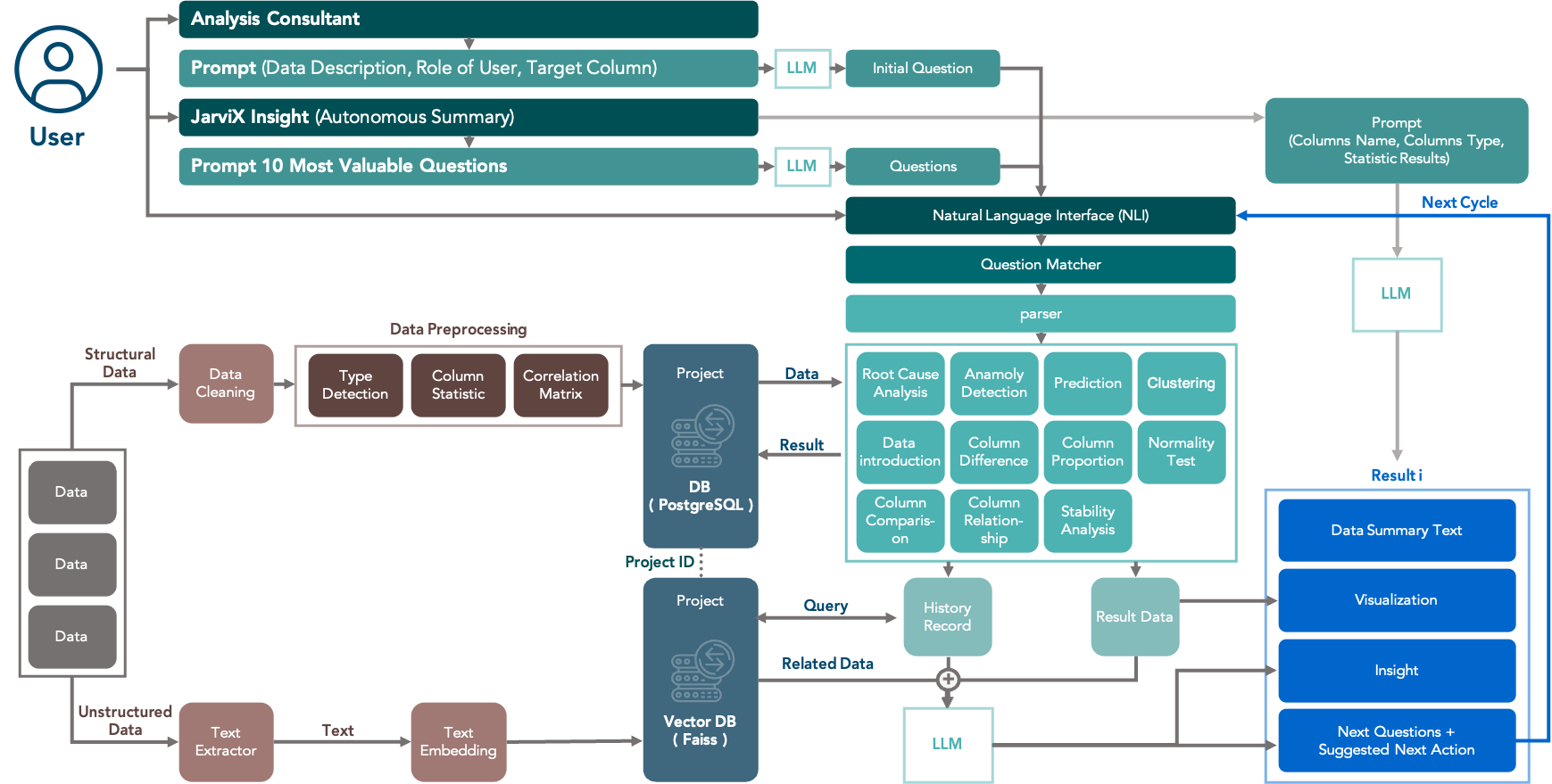}    
    \caption{JarviX system overview}
    \label{fig:mesh1}
\end{figure*}

\section{Introduction}

Although the predominant focus of contemporary research on large language models is the evaluation of various tasks \cite{liang2022holistic, zhao2023survey}, there is a noticeable lack of academic resources that provide structured guidelines and frameworks for downstream applications. 
Tabular data analysis, as an important application task of LLMs, has always faced challenges related to the precision of mathematical calculations. Despite its ability to address complex high school math problems and participate in advanced mathematical discussions, advanced models such as GPT-4 are not yet on par with expert level performance \cite{bubeck2023sparks}. They are prone to basic errors and can sometimes produce incoherent outputs. This may stem from the fact that autoregressive models lack self-correction mechanisms when generating solutions
\cite{shen2021generate}.  
This paper introduces a thorough approach towards employing LLMs for tabular data analysis tasks, specifically aiming to equip nonspecialists with the ability to engage in advanced data analytics using LLMs within a rule-based system. Although LLMs have proven to be potent in data processing \cite{zhao2023survey}, their application in guiding users through rule-based systems to intuitively create data visualizations, synthesize statistical insights, and provide context-aware explanations is significantly underexplored.

LLM guides users through a rule-based system in JarviX, which enables data visualization and statistical analysis. It leverages a vectorized domain knowledge repository to provide relevant explanations for each visualization and suggests further exploration directions \cite{feng2023xnli}. Users can generate additional exploratory charts and insights through text or voice input, facilitated by Whisper.\footnote{https://github.com/openai/whisper} These insights
%sourced from both an unstructured knowledge base and significant data, 
are processed by the Vicuña model\footnote{https://lmsys.org/blog/2023-03-30-vicuna/} through prompts fine-tuned by GPT-4\footnote{https://openai.com/research/gpt-4}. This approach culminates in a comprehensive report encapsulating all the insights and analytical processes, serving as a thorough guide for users and a blueprint for future analysis.
% This process commences with the LLM directing users through a rule-based system, enabling them to generate insightful data visualizations and statistics. Here, the LLM harnesses a vectorized professional domain knowledge repository, allowing it to provide users with relevant explanations for each visualization, as well as suggest potential directions for further exploration similar to XNLI\cite{feng2023xnli}. JarviX, in addition, users can interact with the system either through text or voice input, facilitated by Whisper\cite{whisper}, leading to the generation of additional exploratory charts and insights that comes from both unstructured knowledge base and significant data part and synthesize by GPT-4\cite{bubeck2023sparks} tuned prompt to Vicuña model to get the final result.

% This approach culminates in the generation of a comprehensive report compiling the insights and analysis process, which can be saved for future reference. This method ensures not only that the user comprehends the explored data thoroughly, but also establishes a blueprint for repeated analysis.

Furthermore, this study explores the integration of H2O-AutoML-customized AutoML pipelines \cite{h2o-automl} into this process. The use of AutoML is demonstrated to identify the best targets with respect to other data columns and to build specific models to optimize results for various objectives, such as optimal factory configurations \cite{he2021automl}.

The primary objective of this study is to empower users with the knowledge and tools necessary to harness the power of LLM for rule-based data analytics by fine-tuning \cite{chung2022scaling} and AutoML. The paper concludes by underlining the potential of this approach in democratizing data analytics, thereby fostering more strategic and informed decision-making.
\section{Related Work}
\subsection{Natural Language Interfaces for Data analysis}
Natural Language Interfaces have recently garnered attention and integration into various commercial data analysis and visualization software, such as IBM Watson Analytics \cite{ibm_watson}, Microsoft Power BI \cite{microsoft_powerbi}, Tableau \cite{tableau}, %ThoughtSpot\cite{thoughtspot}, 
and Google Spreadsheet \cite{google_sheet}. Despite initial limitations, such as confining natural language interactions to data queries and standard chart types, the approach is evolving. Current methods of Natural Language Processing (NLP) incorporate heuristic algorithms, rule-based systems, and probabilistic grammar-based approaches, each with their respective challenges and trade-offs in accuracy, flexibility, and computational resources \cite{miwa2016end,voigt-etal-2021-challengeses,satyanarayan2016vega}.

\subsection{Utilization of LLMs in Advanced Data Analysis}
Rajkumar's performance evaluation of LLM on Text2SQL \cite{rajkumar2022evaluating} was a significant development. Despite the considerable progress, including Sun et al.'s 
 \cite{sun2023sqlpalm} Text2SQL method achieving 77.3\% accuracy on the Spider benchmark, constructing a seamless pipeline is still challenging. The optimization strategy of Hu et al. \cite{hu2023chatdb} and the question refinement strategy of Guo et al. \cite{guo2023retrievalaugmented} show further improvements. The Maddigan et al. system \cite{maddigan2023chat2vis} underscores the importance of visualization post-SQL generation. However, the demand for practical solutions for real-world applications remains, prompting the development of the JarviX platform. It bridges the gap, offering higher-level APIs and integrating LLMs for a comprehensive solution for advanced data analysis.

\subsection{External Knowledge Integration}
Currently, LLMs are confronted with two issues: privacy implications and the obsolescence of training data. Utilizing user-interaction data for further training in online LLM applications can potentially jeopardize security. Additionally, the significant costs associated with retraining can make LLMs outdated over time. LangChain \cite{Chase_LangChain_2022} provides an innovative solution that continuously embeds the latest data and retrieves relevant information from its database, consequently generating current responses while preserving privacy. Moreover, the introduction of the llama\_index \cite{liu_llamaIndex_2022} proposes a more structured approach for embedding levels to retrieve query-related information, such as identifying the latest facts or providing relational data. This method improves the precision of LLM responses.

\begin{figure*}[htb!]
    \centering
    \includegraphics[width=\textwidth]{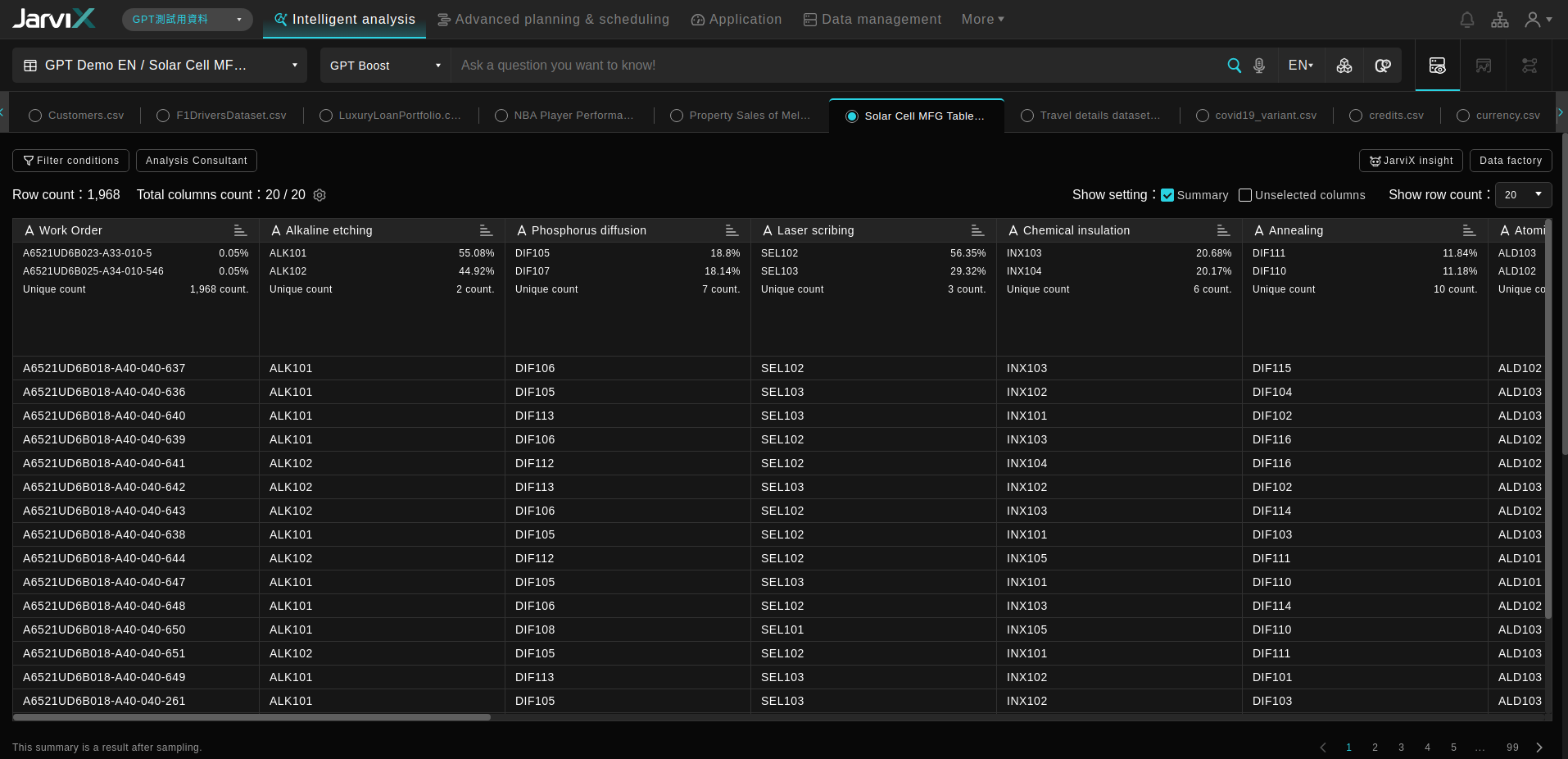}    
    \caption{Main page}
    \label{fig:Case_1_1}
\end{figure*}

\begin{figure}[!hbt]
    \centering
    \includegraphics[width=0.5\textwidth]{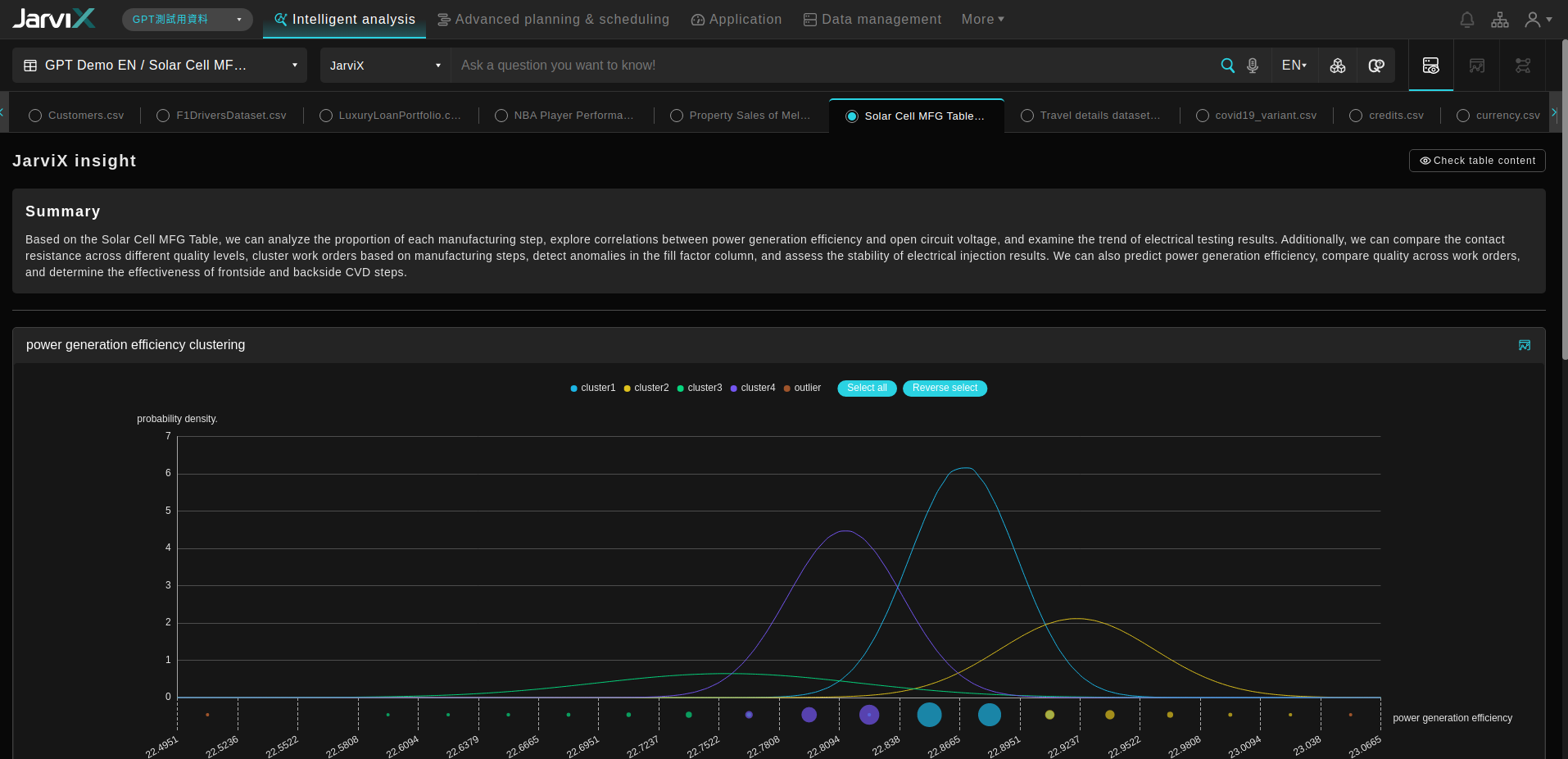}
    \caption{JarviX insight}
    \label{fig:jarvix_insight}
\end{figure}
\section{Overview}
JarviX is a no-code platform for efficient analysis and optimization of tabular data, handling both structured and unstructured types, as illustrated in Figure \ref{fig:mesh1}. For structured data(e.g., csv files, data frames), it performs preliminary processing tasks including data type detection, statistical computation, and correlation analysis, storing the results in a Postgres database. Unstructured data (e.g., text files, audio files) is managed through text extraction and embedding, followed by storage in a vectorized database, such as Elastic Search.
%JarviX is a no-code platform that offers a seamless experience for analyzing and optimizing tabular data. It processes two types of data: structured and unstructured. The platform carries out initial data cleaning and preparation tasks on the structured data, such as column data type detection, statistical computation, and inter-column correlation analysis, which are all stored in a Postgres Database. Unstructured data undergoes text extraction and embedding, which is subsequently saved in a vectorized database, like Elastic Search.

Users can interact with the platform through three key features: JarviX Insight, Natural Language Interfaces, and JarviX Guidance. JarviX Insight collects structured data information such as column names, types, and statistical data, and employs a LLM to generate a data summary report, providing users with an understanding of their data and identifying key questions.

The Natural Language Interfaces feature accommodates user queries about their datasets, either voice-to-text or typed, and translates these queries into a rule-based system via a fine-tuned LLM. This delivers relevant data visualizations, explanations, and follow-up question suggestions.

%The NLI feature allows users to inquire about their datasets through voice-to-text or typed sentences. Leveraging a fine-tuned LLM, the platform translates user questions into a rule-based system, providing appropriate data visualizations, explanations, and suggestions for potential follow-up questions.

Lastly, JarviX Guidance assists users through a step-by-step data analysis process. It takes into account the user's understanding of the datasets, their role, the specific dataset, and target column they wish to analyze. Using this information, JarviX anticipates the questions a user might want to address first and commences the result generation process. It also provides an appropriate endpoint for analysis for each user. All stages of the analysis are recorded, including the middle $result_i$, and compiled into a comprehensive report that users can save and share.

\section{System Break Down}
%\subsection{Any Language to rule based fine-tuning/Query Interpretation}
\subsection{Data Input Methods}
JarviX offers users three methods for uploading structured data: via SFTP, database connections, or direct CSV file uploads. For unstructured data, the platform currently supports only file uploads.

JarviX integrates a data cleaning interface with automated functions, enabling users to efficiently perform data cleaning. Then, the system initiates data pre-processing, which includes automatic type detection, column statistics computation, and calculation of the correlation matrix between columns. It is worth noting that these tasks—column statistical computation and correlation matrix calculation—are executed asynchronously, ensuring that user progress isn't hindered.

In handling unstructured data, we leverage various connectors in the llama hub \cite{Zhang_2023} to perform text extraction. The extracted data is stored in the vector database using Faiss \cite{githubGitHubFacebookresearchfaiss} and assigned the same project ID as the structured database. This cohesive data management strategy ensures seamless integration and retrieval of both structured and unstructured data.

\subsection{JarviX Insight}

The JarviX Insight feature facilitates autonomous report generation, enabling users to comprehend data more effectively and gain insight into potential subsequent questions, as illustrated in Figure \ref{fig:mesh1} and shown in Figure \ref{fig:jarvix_insight}. Upon activation of the JarviX Insight function, two distinct processes occur. Initially, a prompt with preprocessed data is used to determine the nature of the data, which subsequently assists in the creation of a data summary text. Currently, the LLM is used to generate the ten most pertinent questions. These questions serve as a foundation for generating a variety of potential visualization results. By integrating these elements, a comprehensive data summary report is crafted.

\subsection{Question Matcher}

The Question Matcher is the key module that links questions from a natural language interface to their corresponding modules using SQL matching. This process relies on identifying three types of keywords: 1) column name-related terms, 2) restriction-related phrases (e.g. "top ten"), and 3) algorithm or module keywords. Once these keywords are identified, the module begins to merge the specific restrictions associated with each column into a unified combination. This combination is then matched with the algorithm or module indicated by the third type of keyword. More details on Question Matcher are contained in the Appendix \ref{appendix}.


\subsection{Analysis Consultant}

The Analysis Consultant is designed for users who have an initial comprehension of their data and express interest in exploring specific columns, as depicted in Figure \ref{fig:mesh1}. The process begins with the setup of the analysis parameters based on previously outlined criteria. Subsequently, the LLM formulates the first query. The Consultant then generates comprehensive results that include visualizations, insights with supportive explanations, and prompts for potential follow-up queries from the users. A crucial aspect of this process is the incorporation of professional knowledge into the insights, providing not only a fundamental explanation of the visualizations but also integrating general knowledge and background understanding into the explanation. If the analysis process is deemed comprehensive, the Consultant may propose generating a report.

LLMs put forth subsequent analytical queries, facilitating users in delving deeper into their target column's data features. The formulation of questions is based on the preceding results of the analysis and the roles selected by the user. This feature equips people who lack in-depth data analysis knowledge to yield thorough and dependable interpretations of their data. A step-by-step description of this process will be exhibited in Session \ref{sec:case_study}.

\subsection{LLM, Prompt Engineering}
In our experiments, we leverage the vicuna-13b-1.1-gptq-4bit-128g \cite{vicuna-13B-1.1-GPTQ-4bit-128g} as our base LLM, and optimize the prompts with the advanced language understanding capabilities of GPT-4 to achieve better results.

\begin{figure}
    \centering
    \includegraphics[width=0.45\textwidth]{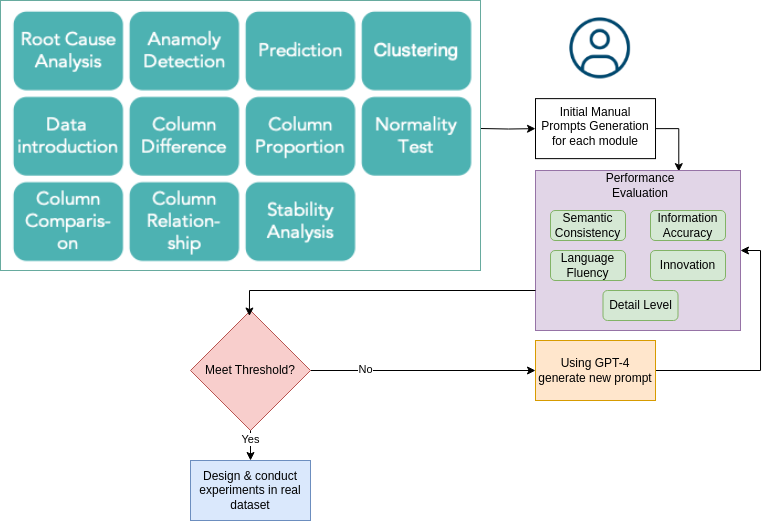}
    \caption{Prompt optimization process}
    \label{fig:prompt optimization process}
\end{figure}

\begin{figure*}
    \centering
    \includegraphics[width=\textwidth]{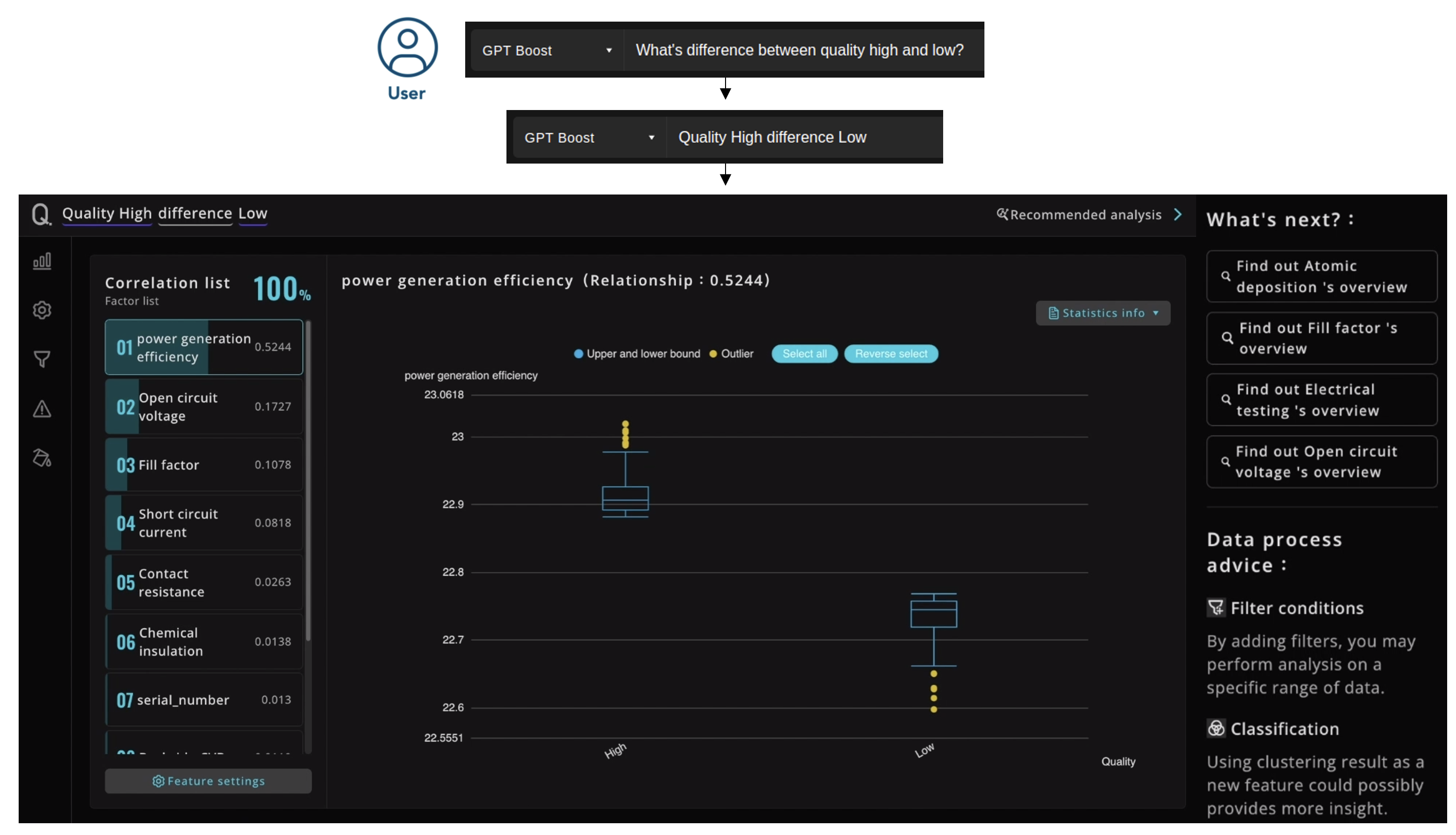}
    \caption{Question matcher}
    \label{fig:Question matchers}
\end{figure*}
% The core of our approach is prompt engineering, which can be understood as a two-fold methodology, as figure \ref{fig:prompt optimization process}. Initially, prompts are generated manually, considering a nuanced understanding of the task module at hand. This necessitates the analysis diagram of the structure along with the task goal, followed by crafting prompts that are both specific and concise. 

Our approach, based on prompt engineering, can be viewed as a two-stage process, as detailed in Figure \ref{fig:prompt optimization process}. First, we manually generate prompts, which require a deep understanding of the task module and a clear view of the structure and goal of the task. The resultant prompts are both specific and concise. 

% After the generation of initial prompts, we apply a feedback-based loop to optimize them. Each prompt is input into the Vicuña model, generating respective outputs. These are compared with the expected results to calculate a performance evaluation metric. If a prompt performs below our threshold, we use GPT-4 to generate a new prompt to replace the original.

Post-initial generation, we instigate a feedback loop to optimize the prompts. Every prompt is fed into the Vicuña model to generate respective outputs, which are compared to the expected results to derive a performance metric. If a prompt falls short of our performance benchmark, it's replaced with a new one generated by GPT-4.

% This process of prompt optimization and replacement continues until all prompts meet our set performance standards. To validate our methodology, we conduct several experiments that span a variety of tasks including normality tests, forecasting, comparisons, root cause analysis, anomaly detection, and relationship extraction.

We persist with this prompt optimization until all prompts meet our performance criteria. To corroborate our approach, we undertake a range of experiments spanning tasks like normality tests, forecasting, comparisons, root cause analysis, anomaly detection, and relationship extraction.
% TODO: add appendix

% Our primary objective through these experiments is to contrast our approach's performance against existing prompt engineering methods. The results, as demonstrated through various examples and illustrations from our experiments, prove the robustness of our methodology. The Vicuña model, interacting with the prompts and feedback loops, generates enriched and insightful responses to complex data-driven queries.

The primary aim of our experiments is to juxtapose the performance of our approach against existing prompt engineering techniques. The results, presented through multiple examples from our experiments, attest to the strength of our methodology. Taking advantage of the interaction between prompts and feedback loops, the Vicuña model yields rich and insightful responses to complex data-related queries.

\section{Case Study}
\label{sec:case_study}
In this section, we demonstrate two separate use cases: 1) Utilizing JarviX Insight for custom analysis with data matching, and 2) Exploring JarviX's guidance use cases.

\subsection{Case 1: JarviX Insight and Solar Cell Manufacturing}

We present a case that explores solar cell manufacturing data and demonstrates how to increase efficiency using JarviX Insight.

As shown in Figure \ref{fig:Case_1_1}, users begin by using JarviX Insight to generate a report for general understanding of the data set. If users are unfamiliar with the data, the JarviX Insight function can provide a general report that answers two questions: 1) What is the subject of these data? 2) What are the most valuable queries that can be made using this dataset?

%JarviX Insight generate a report that addresses these questions, providing a summary of the dataset and visualizations that offer insight into the most valuable queries.

% \begin{figure}[!hbt]
%     \centering
%     \subfloat[\centering Natural language input]{{\includegraphics[width=0.45\textwidth]{img/Case_1_3_1.png} }}%
%     \qquad
%     \subfloat[\centering keywords of JarviX]{{\includegraphics[width=0.35\textwidth]{img/Case_1_3_2_1.png} }}%
%     \qquad
%     \subfloat[\centering Visualization result]{{\includegraphics[width=0.45\textwidth]{img/Case_1_3_2_2.png} }}%
%     \qquad
%     \subfloat[\centering Next action suggestions]{{\includegraphics[width=0.15\textwidth]{img/Case_1_3_2_3.png} }}%
%     \caption{Question Matcher Function}
%     \label{fig:question_matcher}
% \end{figure}

% Case_1_3_0.png

%Informed by JarviX Insight that quality might be an area for improvement, a user can then use the "Question Matching" function to pose general queries like ``What is the difference between high quality and low quality?" As seen in Figure \ref{fig:question_matcher}, the "Question Matcher" maps the input words to corresponding keywords that the system parser can understand.

Upon gaining insights that quality could be a potential area for enhancement, users can utilize the "Question Matching" feature. This function facilitates the formation of general queries, such as ``What is the difference between high quality and low quality'' As illustrated in Figure \ref{fig:Question matchers}, the ``Question Matcher'' successfully translates user inputs into keywords recognizable by a rule-based system.

The visualization results of different algorithms indicate the key differences between high and low quality. The visualization includes straightforward insights, suggested questions, actions, and the main diagram.

% \begin{figure}[!hbt]
%     \centering
%     \subfloat[\centering Training management page]{{\includegraphics[width=0.35\textwidth]{img/Case_1_4.png} }}%
%     \qquad
%     \subfloat[\centering Training settings]{{\includegraphics[width=0.25\textwidth]{img/Case_1_5.png} }}%
%     \qquad
%     \subfloat[\centering Features selection]{{\includegraphics[width=0.45\textwidth]{img/Case_1_6_1.png} }}%
%     \qquad
%     \subfloat[\centering Strategy selection]{{\includegraphics[width=0.25\textwidth]{img/Case_1_6_2.png} }}%
%     \caption{AutoML Pipeline}
%     \label{fig:automl_pipeline}
% \end{figure}

\begin{figure}
    \centering
    \includegraphics[width=0.45\textwidth]{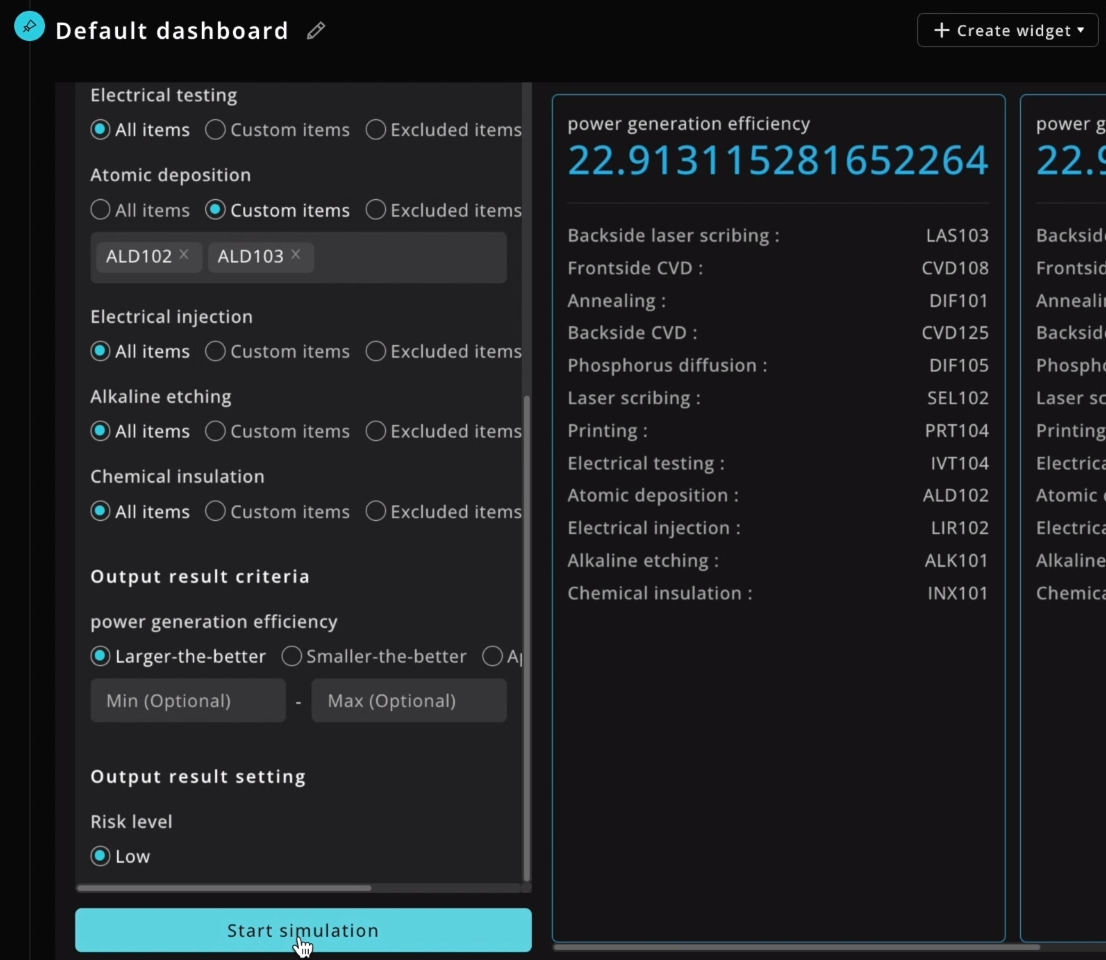}
    \caption{Simulation}
    \label{fig:simulation}
\end{figure}

% The next recommended step is to learn about the best configurations for obtaining high quality results. Our AutoML pipeline, allows users to train a machine learning model by setting the data source, dataset, and target column. You also need to set the features' column, including the performance metric (such as MAE, MSE, or RMSE) you want to optimize. It is also necessary to specify the desired training strategy; the more precise the strategy, the more time it takes to train the model. After completing all these settings, you can generate the model.
Our AutoML pipeline simplifies the process of training a machine learning model. Users simply define the data source, dataset, and target column, as well as the performance metric for optimization (such as MAE, MSE, or RMSE). Though the training strategy's precision may impact the model's training time, once all parameters are set, the model can be generated. Once the model is established, users can explore optimal settings through the simulation panel, as depicted in Figure \ref{fig:simulation}. This panel allows users to identify optimal configurations within the defined range. Importantly, the model is designed to progressively refine itself with the influx of new streaming data. This dynamic adaptation promises improved outcomes over time as settings are intelligently adjusted in response to the evolving data. 

This process provides an ideal optimization cycle for customers. In this case, the solar panel manufacturer increased efficiency by 10\% using this optimization cycle.

\subsection{Case 2: JarviX Guidance, An Analysis of LCD Factory Data}

If users are new to JarviX, understanding its functionality or learning how to analyze data might be challenging. To address this, our analysis consultant is available to guide you through the process. In this case study, our focus is on interpreting a dataframe relevant to an LCD panel factory.

\begin{figure}[!hbt]
    \centering
    \includegraphics[width=0.4\textwidth]{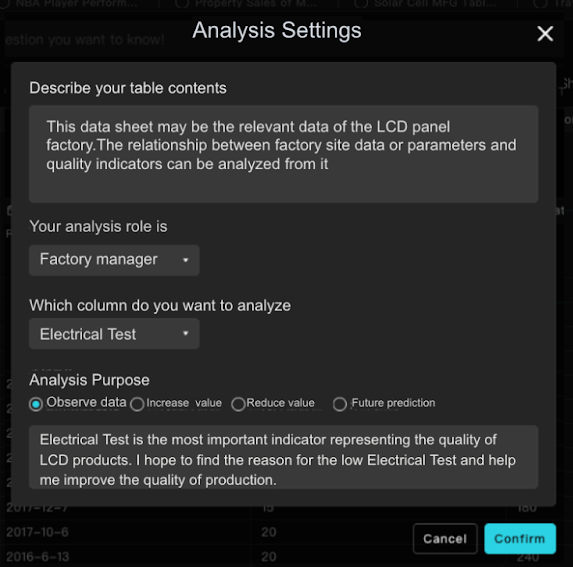}
    \caption{Settings for the analysis}
    \label{fig:AC_2}
\end{figure}

Setting up the system properly is paramount to ensure it identifies the data pertinent to the user. We commence by describing the content of the data table, followed by outlining our analysis objectives and roles, as depicted in Figure \ref{fig:AC_2}.

As an initial step, JarviX recommends the appropriate analysis that users should consider. It's often challenging for users to extract vital information from the analysis process when faced with varied data. Therefore, we inform our model about the analysis approach, so it can recommend suitable subsequent analyses based on users' requirements. 
% Here, we are presented with two options. Let's begin with the first one as trend (Figure \ref{fig:AC_4}).

%\begin{figure}[!hbt]
  %  \centering
    %\includegraphics[width=0.35\textwidth]{img/AC_3.png}
    %\caption{Recommended analyses}
   % \label{fig:AC_3}
%\end{figure}

Upon starting the analysis, JarviX assists users in interpreting the results and guides them to the next steps.
% In figure \ref{fig:AC_4}, part (B) provides insights about the chart, while the part (C), the system suggests potential next steps for users.

% \begin{figure}[!hbt]
%     \centering
%     \subfloat[\centering Visualization]{{\includegraphics[width=0.35\textwidth]{img/AC_4_1.png} }}%
%     \qquad
%     \subfloat[\centering Insight]{{\includegraphics[width=0.2\textwidth]{img/AC_4_2.png} }}%
%     \qquad
%     \subfloat[\centering Next action suggestion]{{\includegraphics[width=0.15\textwidth]{img/AC_4_3.png} }}%
%     \caption{Trend analysis}
%     \label{fig:AC_4}
% \end{figure}

% Guided by JarviX, users can extract valuable information from the analysis process. 

Through our differential analysis, we determined that the electrical test performance heavily depends on the stability of ambient humidity. A list on the left displays the significant factors that influence the differences between high and low electrical tests. In particular, humidity is the top factor, indicating that humidity differences significantly affect the performance of electrical tests.

% \begin{figure}[!hbt]
%     \centering
%     \subfloat[\centering Visualization]{{\includegraphics[width=0.35\textwidth]{img/AC_5_1.png} }}%
%     \qquad
%     \subfloat[\centering Insight]{{\includegraphics[width=0.2\textwidth]{img/AC_5_2.png} }}%
%     \qquad
%     \subfloat[\centering Next action suggestion]{{\includegraphics[width=0.15\textwidth]{img/AC_5_3.png} }}%
%     \caption{Difference analysis}
%     \label{fig:AC_5}
% \end{figure}

\begin{figure}[!hbt]
    \centering
    \subfloat[\centering Summary text]{{\includegraphics[width=0.4\textwidth]{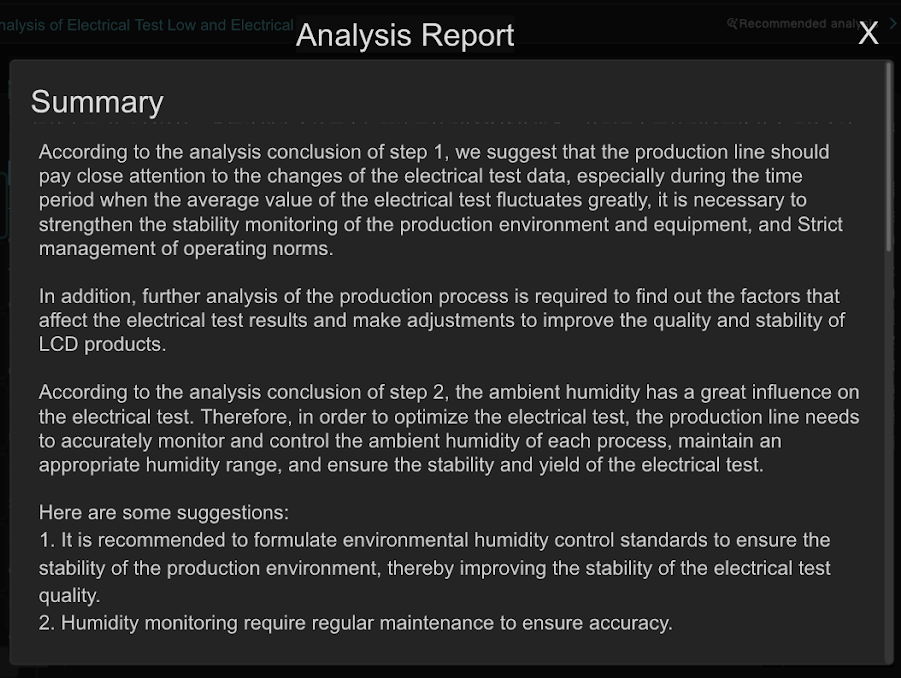} }}%
    \qquad
    \subfloat[\centering Summary figure]{{\includegraphics[width=0.4\textwidth]{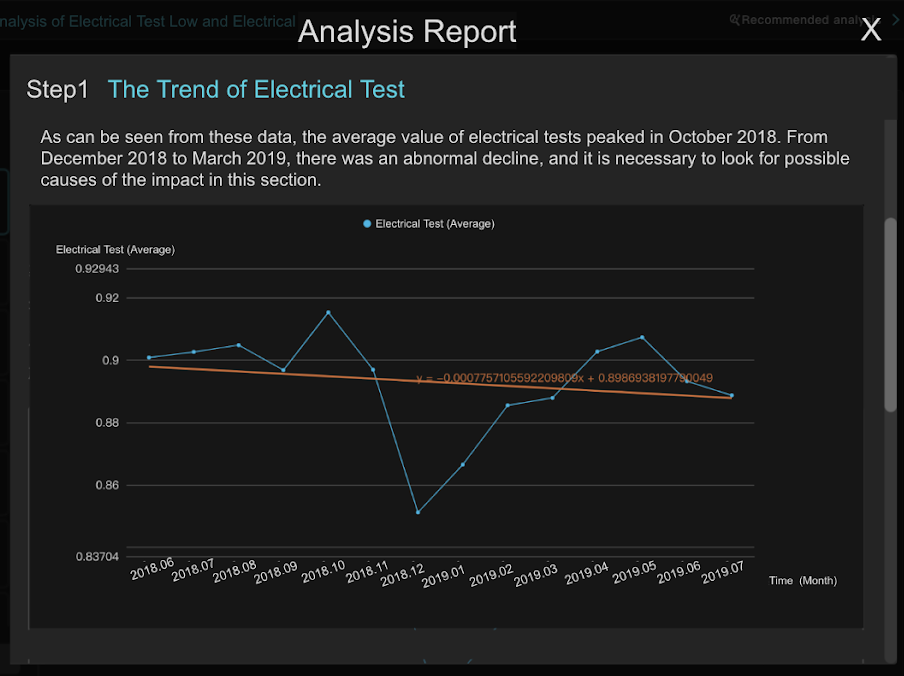} }}%
    \caption{Analysis summary}
    \label{fig:summary}
\end{figure}

At this stage, the system suggests the production of a summary report based on the previous analysis. JarviX will first show a summary suggestion, then recapitulate the previous analysis steps as Figure \ref{fig:summary}. The analysis consultant's guidance enables the user to obtain valuable analysis results, which can help optimize company strategy or uncover potential business value.

\section{Conclusion}
JarviX, by integrating LLM and AutoML technologies, presents a unique and all-encompassing approach for the analysis of tabular data. The system integrates non-structural data to generate profound insights, employing LLM to aid users in their data exploration endeavors.

\section{Future Work}
The flexibility and adaptability of the JarviX platform offer several avenues for future improvements. In particular, there are opportunities to fine-tune the LLM to improve personalized recommendations, extend the range of accepted data types and query categories, and improve user interface design for a better user experience.
\section*{Ethical Considerations and Limitations}
JarviX formulates responses influenced by the user-provided context. Biased results may arise if the context involves biases related to aspects such as the location or language of the user \cite{hadi2023survey}. For example, given that JarviX currently only supports processing and analysis in English and Chinese, it might yield biased answers when inquiries about a specific culture or religion are presented, especially if JarviX lacks adequate training in that particular cultural or religious context, due to its confined knowledge. Also,
JarviX is designed to recognize only plain text information and cannot identify multimodal tabular data, such as financial statements or instructional videos.

\section*{Acknowledgements}
This research is supported by Synergies Intelligent Systems, Inc. and partially supported by the DFG/NSFC Collaborative Project “Crossmodal Learning - Adaptivity, Prediction and Interaction” SFB/TRR169.

% This document has been adapted
% by Steven Bethard, Ryan Cotterell and Rui Yan
% from the instructions for earlier ACL and NAACL proceedings, including those for 
% ACL 2019 by Douwe Kiela and Ivan Vuli\'{c},
% NAACL 2019 by Stephanie Lukin and Alla Roskovskaya, 
% ACL 2018 by Shay Cohen, Kevin Gimpel, and Wei Lu, 
% NAACL 2018 by Margaret Mitchell and Stephanie Lukin,
% Bib\TeX{} suggestions for (NA)ACL 2017/2018 from Jason Eisner,
% ACL 2017 by Dan Gildea and Min-Yen Kan, 
% NAACL 2017 by Margaret Mitchell, 
% ACL 2012 by Maggie Li and Michael White, 
% ACL 2010 by Jing-Shin Chang and Philipp Koehn, 
% ACL 2008 by Johanna D. Moore, Simone Teufel, James Allan, and Sadaoki Furui, 
% ACL 2005 by Hwee Tou Ng and Kemal Oflazer, 
% ACL 2002 by Eugene Charniak and Dekang Lin, 
% and earlier ACL and EACL formats written by several people, including
% John Chen, Henry S. Thompson and Donald Walker.
% Additional elements were taken from the formatting instructions of the \emph{International Joint Conference on Artificial Intelligence} and the \emph{Conference on Computer Vision and Pattern Recognition}.

% test\cite{-casacuberta-2022-shot}
\bibliography{anthology,custom}
\bibliographystyle{acl_natbib}
\clearpage
\appendix

\section{Experiment Results}
\label{appendix}
\subsection{Dataset}
We conducted evaluations on JarviX using a variety of tabular datasets\footnote{https://reurl.cc/jv693q} sourced from open source collections, covering different fields. A set of 10 manually crafted questions was complemented by an additional 20 generated by GPT-4 for each data set. To ensure relevance and meaningfulness, the prompts were designed with the phrase ``assuming that you are a professional data analyst in this field'', tailoring the questions generated to the specific industry. The test data were then classified into six distinct industry-based segments.

\subsection{Results}
The TABLE I  presents the performance results of the question matching evaluation, focusing on three specific aspects: column name, intention, and restriction detection. 
% \begin{itemize}
%     \item Top1: This category indicates the instances where the correct result was identified as the top most (first) result.
%     \item Top3: This category represents the cases where the correct result was found within the top three results, providing a broader view of the system's accuracy.
% \end{itemize}
Each aspect of the evaluation is thoroughly examined, with results meticulously tabulated to offer a comprehensive understanding of the system's performance across the different dimensions. JarviX demonstrates proficient recognition of individual columns. However, when faced with questions that encompass multiple columns, there is a possibility that it might not fully recognize all of them. The eleven intentions that JarviX is capable of executing are illustrated in Figure \ref{fig:prompt optimization process}. In addition, JarviX is equipped to identify the following specific restrictions: {\small \textit{Average}, \textit{Median}, \textit{Sum}, \textit{Greater than}, \textit{Equal to}, \textit{Less than}, \textit{Plus}, \textit{Minus}, \textit{Multiply}, \textit{Divide}, \textit{Top}, \textit{Last}, \textit{Maximum}, \textit{Minimum}}. On the basis of our experimental findings, JarviX shows enhanced performance as user queries exhibit clearer intent. However, when faced with ambiguous queries, JarviX is prone to over-identifying or under-identifying terms.

\begin{table}[t]
\centering
\parbox{\columnwidth}{
\small
    TABLE I : Evaluation result for Question Matching. Top1 indicates the instances where the correct result was identified as the top most result. Top3 represents the cases where the correct result was found within the top three results.
    \par\vspace{5pt} 
}
% \begin{adjustbox}{max width=\textwidth}
\resizebox{\columnwidth}{!}{%
\begin{tabularx}{9.85cm} {cllllll}
\hline
\multirow{2}{*}{Data source} & \multicolumn{2}{c}{Column Name}                       & \multicolumn{2}{c}{Intention}                         & \multicolumn{2}{c}{Restriction}                       
\\ \cline{2-7}  
% &
 & \multicolumn{1}{c}{Top1} & \multicolumn{1}{c}{Top3} & \multicolumn{1}{c}{Top1} & \multicolumn{1}{c}{Top3} & \multicolumn{1}{c}{Top1} & \multicolumn{1}{c}{Top3} \\ \hline
Manufacture                  & 72.0                             & 83.3                             & 74.0                             & 82.7                             & 64.7                             & 72.0                             \\
Sport                        & 73.3                             & 88.3                             & 75.0                             & 90.8                             & 65.8                             & 76.7                             \\
Sales                        & 70.7                             & 82.7                             & 77.3                             & 86.7                             & 67.3                             & 78.7                             \\
Food                         & 69.2                             & 88.3                             & 75.8                             & 90.8                             & 70.0                             & 77.5                             \\
Health Care                  & 65.0                             & 74.2                             & 73.3                             & 85.0                             & 67.5                             & 74.2                             \\
Banking                      & 81.7                             & 93.3                             & 79.2                             & 91.2                             & 66.7                             & 74.2                             \\ \hline
\end{tabularx}%
}
% \end{adjustbox}
 % \caption{TABLE I : Evaluation result for Question Matching}
\label{tab:question matching}
\end{table}

\end{document}